# Single Model Deep Learning on Imbalanced Small Datasets for Skin Lesion Classification

Peng Yao, Shuwei Shen, Mengjuan Xu, Peng Liu, Fan Zhang, Jinyu Xing, Pengfei Shao, Benjamin Kaffenberger, and Ronald X. Xu

***Abstract*—Deep convolutional neural network (DCNN) models have been widely explored for skin disease diagnosis and some of them have achieved the diagnostic outcomes comparable or even superior to those of dermatologists. However, broad implementation of DCNN in skin disease detection is hindered by small size and data imbalance of the publically accessible skin lesion datasets. This paper proposes a novel single-model based strategy for classification of skin lesions on small and imbalanced datasets. First, various DCNNs are trained on different small and imbalanced datasets to verify that the models with moderate complexity outperform the larger models. Second, regularization DropOut and DropBlock are added to reduce overfitting and a Modified RandAugment augmentation strategy is proposed to deal with the defects of sample underrepresentation in the small dataset. Finally, a novel Multi-Weighted New Loss (MWNL) function and an end-to-end cumulative learning strategy (CLS) are introduced to overcome the challenge of uneven sample size and classification difficulty and to reduce the impact of abnormal samples on training. By combining Modified RandAugment, MWNL and CLS, our single DCNN model method achieved the classification accuracy comparable or superior to those of multiple ensembling models on different dermoscopic image datasets. Our study shows that this method is able to achieve a high classification performance at a low cost of computational resources and inference time, potentially suitable to implement in mobile devices for automated screening of skin lesions and many other malignancies in low resource settings.**

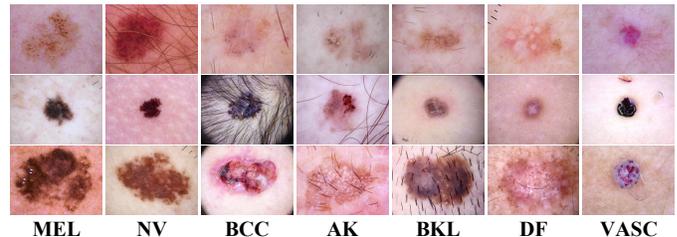

Fig. 1. Representative images of seven lesion classes in the HAM10000 dataset [2] that show variations between low inter-class and high intra-class lesions, artifacts, relatively low contrast and fuzzy borders between lesions and normal skin areas.

## I. INTRODUCTION

SKIN cancer is one of the most common malignancies in the world with significantly increased incidence over the past decade [1]. Skin cancer is typically diagnosed based on dermatologists' visual inspection, with support of dermoscopic imaging and confirmation of skin biopsy [3]. However, owing to the shortage of dermatologic work force and the lack of pathology lab facility in rural clinics, patients in rural communities do not have access to prompt detection of skin cancer, leading to the increased morbidities and melanoma mortalities [4]. Artificial intelligence based on deep learning (DL) represents a major breakthrough for computer-aided diagnosis of cancers [5]. Applying DL techniques in skin lesion classification may potentially automate the screening and early detection of skin cancer despite the shortage of dermatologists and lab facilities in rural communities [6].

Earlier approaches for computer-aided dermoscopic image analysis rely on extraction of handcrafted features to be fed into conventional classifiers [7, 8]. Recently, automatic skin cancer classification has significantly improved the performance by using end-to-end training of deep convolutional neural networks (DCNN) [9-13]. Despite these promising research progresses, further improvement of the diagnostic accuracy is hindered by several limitations. *First of all, most of publicly accessible datasets for skin lesions do not have a sufficient sample size.* Especially, in the scenario where the skin lesions have low contrast, fuzzy borders and interferences such as hair, veins or ruler marks as shown in Fig. 1, a sufficient sample size is the key for appropriate training of a DCNN model to fit the unknown data features, as exemplified by tens of millions of data for algorithm performance verification in the most commonly used ImageNet [14]. Benefited from the large training datasets, previous DCNN models have achieved significant diagnostic outcomes comparable to those of professional dermatologists [12, 13]. However, most of the training datasets for these models are derived from private datasets that are not publicly accessible. In contrast, many publicly available dermoscopic image datasets generally have only a few thousand samples. For example, the HAM10000

Peng Yao and Shuwei Shen contributed equally to this work. Corresponding author: Ronald X. Xu.

P. Yao, M. Xu, P. Liu, F. Zhang, J. Xing, P. Shao, and R. Xu are with the Key Lab Precis Sci Instrument at Anhui Higher Education, University of Science and Technology of China, Hefei 230027, China (e-mail: yaopeng@ustc.edu.cn; xumj@mail.ustc.edu.cn; lpeng01@ustc.edu.cn; zxt1002@mail.ustc.edu.cn; jinyux@ustc.edu.cn; spf@ustc.edu.cn; xux@ustc.edu.cn).

S. Shen and R. Xu are with the Suzhou Advanced Research Institute, University of Science and Technology of China, Suzhou 215000, China (e-mail: swshen@ustc.edu.cn).

B. Kaffenberger is with Division of Dermatology at The Ohio State University Medical Center, Columbus 43210, USA (e-mail: benjamin.kaffenberger@osumc.edu).

TABLE I
CLASS DISTRIBUTION OF HAM AND ISIC 2019 TRAINING SET. THE LESION TYPES ARE MELANOMA (MEL), MELANOCYTIC NEVUS (NV), BASAL CELL CARCINOMA (BCC), ACTINIC KERATOSIS (AK), BENIGN KERATOSIS (BKL), DERMATOFIBROMA (DF), VASCULAR LESIONS (VASC) AND SQUAMOUS CELL CARCINOMA (SCC).

|          | MEL  | NV    | BCC  | AK  | BKL  | DF  | VASC | SCC |
|----------|------|-------|------|-----|------|-----|------|-----|
| HAM      | 1113 | 6705  | 514  | 327 | 1099 | 115 | 142  | 0   |
| ISIC 2019| 4522 | 12875 | 3323 | 867 | 2624 | 239 | 253  | 628 |

dataset (HAM), one of the most commonly used dermoscopic image datasets, consists of only 10015 images [2]. Even the BCN_20000 dataset, one of the largest dermoscopic image datasets, contains 19424 images [2, 15]. Considering the accuracy and stability requirements in diagnostic applications, it is preferred that DCNN models are trained on datasets of the highest possible quality. However, accumulating a large number of clinical data with high quality and consistency in a short time is sometimes difficult owing to many limitations, such as the rarity of the diseases, the data security and privacy concerns, and the lack of medical expertise for appropriate labeling. Given a limited number of available clinical data, it is important to improve the DCNN models trained on small datasets in order to achieve the performance close to that trained on large datasets. *Second, almost all the publicly accessible skin disease image datasets suffer a problem of severe data imbalance.* Since different types of skin lesions have different incident rates and imaging accessibilities, samples among different disease categories typically have uneven distributions, as shown in Table I [2, 15, 16]. Furthermore, many skin lesion images have low inter-class and high intra-class variations, as shown in Fig. 1 [17, 18]. These factors contribute to an imbalanced dataset and poor DCNN performance, especially for rare and similar lesion types. Therefore, it is necessary to optimize the performance of DCNNs for accurate classification of skin lesions regardless of the dataset limitations.

For classification tasks on large-scale image datasets, improving the model structure from initial AlexNet [19] to RegNet [20] or increasing the parameter capacity of the similar model structures [20-22] can always achieve a better performance. However, this is not always true on small image datasets since the increased parameter capacity in this case may induce transition of the models from an under-fitting area to an over-fitting area [23]. Therefore, our first set of scientific questions is: *which model structure yields the best performance and what capacity of networks in similar structures is most suitable for a small dermoscopic image dataset?* Most of the previous researchers select DCNNs without detailing the scientific rationales [12, 13, 24-27]. Yu et al [17] found that a 50-layer ResNet [22] has a satisfactory performance superior to 101-layer and 38-layer ones on a dermoscopic image dataset, but they did not compare the performance variations between different DCNNs. In contrast, Santos et al [28], reported the performance differences in classification tasks for MobileNet [29], VGG-19 [30], and ResNet50 [22], but did not compare the networks of the same series. Our previous research has revealed that EfficientNet-B2 has the best classification performance superior to other series of more capacity and that the increased network complexity may not be suitable for a small dermoscopic dataset [31]. Generally speaking, a DCNN model that performs better on large natural image datasets (e.g. ImageNet) typically performs better in medical image classification tasks. However, it is commonly observed that a moderate model outperforms a larger one in the case of small image datasets. Although no theoretical explanation is available for the above observation yet, we experimentally compare the classification performance of DCNNs with different structures and capacities on the datasets, such as HAM, to verify their feasibility.

In parallel with the search for the most suitable DCNN model, we also need to overcome the intrinsic limitations of the training dataset, such as the small sample size and the image artifacts that make the model prone to overfitting [32]. The second question we focus on in this paper is: *how to reduce overfitting of the DCNN model on the skin lesion dataset that is small and has imaging artifacts?* Generally speaking, the commonly used regularization methods such as DropOut [33] can effectively alleviate overfitting [34, 35]. Ghiasi et al. have demonstrated that DropOut only works in the fully connected layer and that DropBlock works in the convolutional layer with a contribution similar to DropOut [36]. In addition to the above methods, various data augmentation strategies, such as AutoAugment and RandAugment, have also been intensively studied in recent years as an alternative approach to reduce overfitting [37-39]. In comparison with AutoAugment, RandAugment achieves better flexibility and performance at the lower cost of training time and computational resources. In this paper, we explore two approaches to reduce overfitting. First, a new DCNN model is introduced by integrating DropOut and DropBlock. Second, we propose a modified RandAugment strategy more suitable for the limited dermoscopic image data.

In addition to model structure and data augmentation, this paper also attempts to address the third question associated with classification tasks on small image datasets: *how to process the severely imbalanced class in a skin lesion dataset?* Most commonly, the term of "imbalanced class" refers to the imbalanced distribution of sample numbers among different classes, as evidenced in the HAM and ISIC 2019 datasets (Table I). This type of datasets can be processed by sample-based [40, 41] and cost-sensitive-based [42, 43] strategies. The sample-based strategy adds redundant noise data or removes informative training samples, and usually performs less effective than that of the cost-sensitive-based strategy on dermoscopic image datasets [9, 44]. Apart from the imbalanced distribution of sample numbers, the term of "imbalanced class" also refers to the imbalanced classification difficulty between different classes. Lin et al. introduced a Focal Loss method to enhance the train outcome on samples with imbalanced classification difficulties [45]. Cui et al. further proposed a Class-Balanced Focal Loss function that combines the Focal Loss with the Class-Balanced Loss [46]. Considering that accurate detection of melanoma has the highest clinical impact, we propose a Multi-weighted New Loss (MWNL) method that not only overcomes the class imbalance issue in sample number and classification difficulty, but also improve the accuracy of melanoma classification by adjusting the weight of the corresponding loss.

On the other hand, the random cropping expanding the scale of the dataset is widely used in the training of the DCNNs, and it can also greatly enhance the spatial robustness of the model [47]. Noticeably, the skin lesions on some images are very small, and the random cropped images may contain only partial or even empty lesions. Such samples, as well as those with incorrect labels, are called "very hard examples" or "outliers". They may still bring large losses in the convergence training stage. Thereafter, if the converged model is forced to learn to classify these outliers better, the classification of a large number of other examples tends to be less accurate [48]. To deal with this problem, we further introduce a correction operation in our MWNL method that forcibly limits these very large losses in order to reduce the interference of outliers in the network training.

Moreover, Zhou et al. [49] observed that the conventional class re-balancing methods may cause the unexpected loss of the representative ability for the learned deep features, despite their significant promotion of classifier learning. Therefore, they proposed a unified Bilateral-Branch Network (BBN) model to balance between representation learning and classifier learning [49]. Similarly, Cao et al. [50] separated the training process into two stages, where they first trained networks as usual on the originally imbalanced data, and only utilized re-balancing at the second stage to fine-tune the network at a smaller learning rate. Inspired by these previous works, we propose a novel end-to-end cumulative learning strategy capable of effective balancing between representation learning and classifier learning at no additional computational cost. This strategy includes two phases. In the initial phase, the conventional training is carried out on the originally imbalanced data to initialize appropriate weights for deep layers' features [50]. As the number of iterations increases, the training gradually changes to a re-balancing mode, and the learning rate also synchronously decreases to promote the optimization of upper classifier of DCNN.

Although previous studies show that the ensemble DCNN models usually achieve better performance [44, 51-55], implementing these models in a mobile device is not practical, especially in remote and rural areas with limited computational resource. Also, cloud-based intelligent diagnosis is only implementable in developed countries or areas where advanced infrastructure has been established. Considering the urgent need for portable, low-cost, and automatic diagnosis at a low computational cost, we focus our research on the single-model-based method.

The main contributions of this paper are thus summarized as follows:

1) We propose a novel Multi-weighted New Loss method to deal with class imbalance issue and to improve the accuracy for detection of key classes such as melanoma. A correction operation that forcibly limits very large losses is also introduced to reduce the interference of outliers in the network training. As far as we know, the Multi-weighted New Loss method is the first reported that can simultaneously deal with the data imbalance problem and the outlier problem.
2) We propose a novel end-to-end cumulative learning strategy that can balance representation learning and classifier learning more effectively at no additional computational cost. This strategy is designed to first learn in the universal pattern that lead to a good initialization and then gradually focus on the imbalance data.
3) By comparing the classification performance of DCNNs with different structures and capacities on the dermoscopic image datasets, we experimentally verify that the advanced DCNN models performing better on large natural image datasets (e.g. ImageNet) will generally have better performance in medical image classification tasks, and for small image datasets, a moderate model, instead of the larger ones, should yield better performance. To the best of our knowledge, this is the first systematic verification on skin lesion datasets and the first adoption of RegNets in dermoscopic image classification tasks.
4) In addition to the fundamental innovations as listed above, we also improve the available methods for better performance. To increase data diversity of dermoscopic image datasets, we propose a novel Modified RandAugment strategy. To reduce over-fitting, we redesign the DCNN models by integrating regularization DropOut and DropBlock.

We have implemented the proposed methods to the following four datasets: ISIC 2018 [56], ISIC 2019 [2, 15, 16], ISIC 2017 [16], and Seven-Point Criteria Evaluation (7-PT) Dataset [57]. Our experiments have achieved the outstanding performance on all of these datasets. Our next plan is to implement the proposed strategies in a mobile dermoscopic device for low-cost automated screening of skin diseases in low-resource settings.

The impact of our study is not limited to skin disease classification tasks. The methods described in this paper can be further extended to handle the common problems, such as insufficient sample size, class imbalance, and labeling noise, in other medical image datasets and computer vision (CV) classification tasks in general. Although previous publications have addressed each of these problems in depth, few work has been published on simultaneous handling of them. All the source codes of our methods are available at https://github.com/yaopengUSTC/mbit-skin-cancer.git.

## II. RELATED WORK

### A. DCNN Models

Recent research and development efforts on advanced DCNN models have facilitated automated feature extraction and classification with high performance [19-22, 30, 58, 59]. He et al. introduced shortcut connections in ResNet [22] to address the degradation problem [60], making it possible to train up to hundreds or even thousands of layers. ResNet won the 2015 ImageNet image classification competition [14]. The SENet model enhanced its sensitivity to channel characteristics by introducing Squeeze-and-Excitation (SE) [58]. SENet won the 2017 ImageNet image classification competition. In 2019, Tan et al. proposed a series of lightweight DCNN models named EfficientNets [21]. Owing to their outstanding performance in many classification tasks, they have been adopted as a backbone in the latest skin lesion classification tasks [55, 61, 62]. Recently, Radosavovic et al. proposed RegNets that perform better and up to 5x faster on GPUs compared with EfficientNets of similar FLOPs and parameters [20]. RegNets consist of two sequences of RegNetXs and RegNetYs. RegNetXs were designed mainly on standard residual bottleneck block with grouped convolution, while RegNetYs were optimized by the addition of SE modules to RegNetXs [58]. In each of these series, the author trained DCNNs of FLOPs from 200M to 32G on ImageNet, and the performances of RegNetYs were better than those of RegNetXs, proving the effectiveness of the SE structure module for natural image classification.

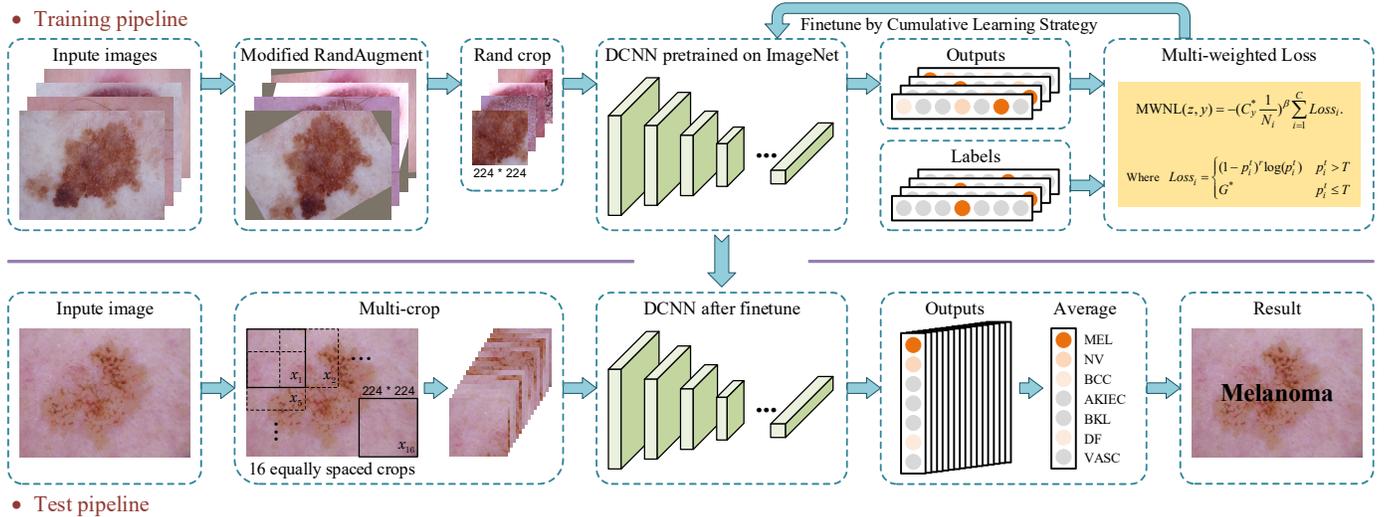

Fig. 2. The flowchart of the proposed framework. The upper part is the training pipeline, and the lower part is the test pipeline.

### B. Data Augmentation

Data augmentation is one of the essential keys to overcome the challenge of limited training data by randomly "augment" data diversity and number. Since the design of augmentation policies requires expertise to capture prior knowledge in each domain, it is difficult to extend existing data augmentation methods to other applications and domains. The recently emerging automatic design augmentation methods based on Learning policies is able to address the shortcomings of traditional data augmentation methods [37-39]. Among them, the RandAugment method proposed by Cubuk et al. is the most advanced data augmentation technology so far [39]. Compared with AutoAugment [37] and Fast AutoAugment [38], the RandAugment greatly reduces the search space and thereby shorten the training time and the computational cost as well.

The search space of RandAugment consists of 14 available transformations. For transformation of each image, a parameter-free procedure is applied in order to reduce the parameter space while maintain the image diversity. RandAugment comprises 2 integer hyperparameters $N$ and $M$, where $N$ is the number of transformations applied to a training image, and $M$ is the magnitude of each augmentation distortion. A randomly selected transformation is applied to each image according to the preset magnitude, followed by repetition of this process for $N$-1 times. All the transformations use the same global parameter $M$ so that the resultant search space size is significantly reduced from $10^{32}$ of AutoAugment [37] and Fast AutoAugment [38] to $10^2$.

### C. Class-Balanced Loss and Focal Loss

To address the class imbalance issue in a given dataset, the cost-sensitive-based methods are commonly used. The methods usually introduce a loss weighting factor inversely proportional to the number of samples [44]:

$$w_i = \frac{1}{N_i}. \tag{1}$$

where $w_i$ is the loss weighting factor for class $i$, and $N_i$ is the number of samples for class $i$.

Cui et al. design a new re-weighting scheme that uses the effective number of samples for each class to re-balance the loss in Class-Balanced Loss [46]. The effective number of samples is defined as $(1-\beta^n)/(1-\beta)$, where $n$ is the number of samples and $\beta \in [0:1)$ is a hyperparameter. The loss weighting factor $w_i$ for class $i$ is thus defined as the class-balanced loss in expression (2):

$$w_i = \frac{1-\beta}{1-\beta^{N_i}}. \tag{2}$$

Notably, $w_i$ varies in [1, $1/N_i$) following the change of $\beta$. As $\beta \to 1$, $w_i \to 1/N_i$. Therefore, we can finally find an optimal $\beta$ value to minimize the performance loss caused by unbalanced samples among classes for any dataset.

Alternatively, Lin et al. applies a modulating term to cross-entropy loss in order to enhance the train outcome on

TABLE II
DETAILED INFORMATION OF DCNNS WITH DIFFERENT PARAMETERS AND DIFFERENT ARCHITECTURES [20-22, 30, 58, 59].

| model | params (M) | flops (B) | model | params (M) | flops (B) | model | params (M) | flops (B) |
|---|---|---|---|---|---|---|---|---|
| VGG-11 | 132.9 | 7.6 | DenseNet-169 | 14.2 | 3.4 | RegNetX-3.2G | 15.3 | 3.2 |
| VGG-13 | 133.0 | 11.3 | DenseNet-201 | 20.0 | 4.3 | RegNetX-4.0G | 22.1 | 4.0 |
| VGG-16 | 138.3 | 15.5 | DenseNet-161 | 28.9 | 7.8 | RegNetX-8.0G | 39.6 | 8.0 |
| VGG-19 | 143.65 | 19.6 | EfficientNet-b0 | 5.3 | 0.39 | RegNetX-16G | 54.3 | 15.9 |
| ResNet-18 | 11.7 | 1.8 | EfficientNet-b1 | 7.8 | 0.7 | RegNetX-32G | 107.8 | 31.7 |
| ResNet-34 | 21.8 | 3.7 | EfficientNet-b2 | 9.2 | 1.0 | RegNetY-400M | 4.3 | 0.4 |
| ResNet-50 | 25.6 | 4.1 | EfficientNet-b3 | 12 | 1.8 | RegNetY-800M | 6.3 | 0.8 |
| ResNet-101 | 44.6 | 7.8 | EfficientNet-b4 | 19 | 4.2 | RegNetY-1.6G | 11.2 | 1.6 |
| ResNet-152 | 60.2 | 11.5 | EfficientNet-b5 | 30 | 9.9 | RegNetY-3.2G | 19.4 | 3.2 |
| SENet-50 | 28.1 | 4.3 | EfficientNet-b6 | 43 | 19 | RegNetY-4.0G | 20.6 | 4.0 |
| SENet-101 | 49.3 | 7.6 | EfficientNet-b7 | 66 | 37 | RegNetY-8.0G | 39.2 | 8.0 |
| SENet-152 | 66.8 | 11.3 | RegNetX-400M | 5.2 | 0.4 | RegNetY-16G | 83.6 | 15.9 |
| SENet-154 | 114.3 | 22.6 | RegNetX-800M | 7.3 | 0.8 | RegNetY-32G | 145.0 | 32.3 |
| DenseNet-121 | 8.0 | 2.9 | RegNetX-1.6G | 9.2 | 1.6 | | | |

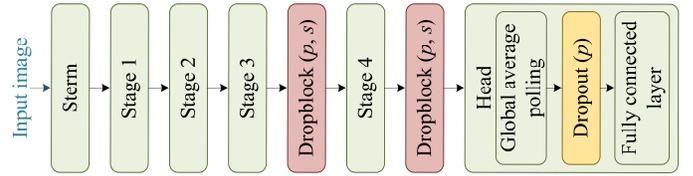

Fig. 3. The architecture of RegNetY-##-Drop

samples with imbalanced classification difficulties [45] [46]. First, a parameter $z_i^t$ is defined as:

$$z_i^t = \begin{cases} z_i, & \text{if } i = y. \\ -z_i, & \text{otherwise.} \end{cases} \quad (3)$$

where $z_i$ is the predicted value of the sample for class $i$, $y$ is the ground truth. Denote $p_i^t = \text{sigmoid}(z_i^t) = 1/(1+\exp(-z_i^t))$, the Focal Loss is then updated as:

$$\text{FL}(z, y) = -\sum_{i=1}^{C}(1-p_i^t)^r \log(p_i^t). \quad (4)$$

where the parameter $r$ smoothly adjusts the rate at which easy examples are down-weighted. As $r = 0$, the Focal Loss value is equivalent to cross entropy (CE) loss. As $r$ increases, the effect of the modulating factor increases accordingly. Based on the Class-Balanced Loss and the Focal Loss, Cui et al. further propose the Class-Balanced Focal Loss ($\text{CB}_{\text{focal}}$) that is able to deal with the imbalances in both the sample numbers and the classification difficulties [46]. The $\text{CB}_{\text{focal}}$ is expressed as following:

$$\text{CB}_{\text{focal}}(z, y) = -\frac{1-\beta}{1-\beta^{N_y}}\sum_{i=1}^{C}(1-p_i^t)^r \log(p_i^t). \quad (5)$$

where $N_y$ is the number of samples in the ground-truth class $y$.

## III. METHODOLOGY

The proposed classification framework is illustrated by the flowchart in Fig. 2. This section first introduces the datasets and the evaluation metrics used in our study, followed by the in-depth explanation of the proposed approaches, including DCNN models, Modified RandAugment, Multi-weighted New Loss method and Cumulative Learning Strategy. Finally, the training and the evaluation strategies are introduced.

### A. Datasets

**ISIC 2018.** ISIC 2018 skin lesion classification challenge [56] adopted the HAM10000 dataset (HAM) [2] as training dataset. HAM dataset is one of the largest and mostly used skin image datasets publicly available in ISIC archive.[1] It consists of 10,015 skin lesion images in seven skin lesion types, namely malignant melanoma (MEL), melanocytic nevus (NV), basal cell carcinoma (BCC), actinic keratosis/bowen's diseases (AK), benign keratosis-like lesion (BKL), dermatofibroma (DF) and vascular (VASC). The number of images in each class refers to Table I. The test set comprises 1512 skin lesion images without published labels. The only method for performance evaluation is to upload the predicted results to the ISIC website.[2]

**ISIC 2017.** The classification challenge dataset of ISIC 2017 splits into training (n=2000), validation (n=150), and test (n=600) datasets. All the images belong to one of 3 categories, including MEL (374 training, 30 validations, and 117 test), seborrheic keratosis (SK) (254, 42, and 90), and NV (1372, 78, and 393).

**ISIC 2019.** ISIC 2019 dataset is made up of HAM dataset [2], MSK dataset [16] and BCN_20000 dataset [15]. There are 25331 images come from MEL, NV, BCC, AK, BKL, DF, VASC and squamous cell carcinoma (SCC). The number of images in each class in the training set refers to Table I. The labels for 8238 test images are also unpublished, and it is also necessary to upload the predicted results to the ISIC website for performance evaluation.

**7-PT Dataset.** 7-PT Dataset [57] contains 1011 cases (we define each unique lesion as a case) and there is a dermoscopic image and some meta data for each case. Moreover, there is also a corresponding clinical image for each case except for 4 cases. All cases belong to one of 5 categories, including BCC (42), NV (575), MEL (252), MISC (97) and SK (45). Following [57], 413, 203, 395 cases are used as training, validate and test data for fair comparisons. Furthermore, unlike [57], only dermoscopic image data is used here.

### B. Evaluation Metrics

In this study, the balanced accuracy (BACC) is used as the main evaluation measure, as suggested by the ISIC 2018 and ISIC 2019 skin lesion classification challenge [56]. BACC is equivalent to the average sensitivity or recall, which treats all the classes equally, and is expressed as:

$$BACC = \frac{1}{C}\sum_{i=1}^{C}\frac{TP_i}{TP_i + FN_i}. \quad (6)$$

where $TP$ denotes true positives, $FN$ denotes false negatives and $C$ denotes the number of classes.

The averaged specificity and the average area under the receiver operating characteristic curve (AUC) are also reported for the evaluation between results of state-of-the-art algorithms. Particularly, the sensitivity of MEL is separately listed for evaluating the classification performance of life-threatening melanoma.

---

[1] https://isic-archive.com/

[2] https://challenge.isic-archive.com

TABLE III
LIST OF ALL TRANSFORMATIONS CAN BE CHOSEN DURING THE SEARCH.

| | Operation Name | Description | Range of magnitudes |
|---|---|---|---|
| {color} | Sample_pairing [63] | Linearly add the image with another image (selected at random from the same dataset) with weight *magnitude*. | [0, 0.2] |
| | Gauss_noise | Add random Gaussian noise to the image with weight *magnitude*. | [0, 0.2] |
| | Saturation | Adjust the saturation. A *magnitude*=0 gives a gray image, whereas *magnitude*=1 gives the original image. | [0.6, 1.4] |
| | Contrast | Control the contrast. A *magnitude*=0 gives a uniform gray image, whereas *magnitude*=1 gives the original image. | [0.6, 1.4] |
| | Brightness | Adjust the brightness. A *magnitude*=0 gives a black image, whereas *magnitude*=1 gives the original image. | [0.6, 1.4] |
| | Sharpness | Adjust the sharpness. A *magnitude*=0 gives a blurred image, whereas *magnitude*=1 gives the original image. | [0.6, 1.4] |
| | Color_casting [64] | Randomly select a channel in RGB color spaces and add an offset value of *magnitude*. | [-30, 30] |
| | Equalize | Equalize the histogram of each RGB channels of image respectively | |
| | Equalize_yuv | Equalize the histogram of Y channels after transferring image into YUV color spaces | |
| | Posterize | Reduce the number of bits for each pixel to *magnitude* bits. | [0, 3] |
| | AutoContrast | Maximize the image contrast, by making the darkest pixel black and lightest pixel white. | |
| | Solarize | Invert all pixels above a threshold value of *magnitude*. | [128, 255] |
| | Vignetting [64] | Make the periphery of the image dark compared to the image center with rate *magnitude*. | [0.0, 0.6] |
| {shape} | Rotate | Rotate the image *magnitude* degrees. | [-40, 40] |
| | Flip | Flip image randomly in Horizontal and Vertical. | |
| | ShearX(Y) | Shear the image along the horizontal (vertical) axis with *magnitude* degrees. | [-15, 15] |
| | Distortion [64] | Distort the image with *magnitude* degrees. | [0.0, 0.6] |
| | Scale | Scale the image horizontally and vertically with equal *magnitude* degrees | [0.8, 1.2] |
| | Scale_diff | Scale the image horizontally and vertically with different *magnitude* degrees. | [0.8, 1.2] |
| | Cutout [65] | Set a random square patch of side-length *magnitude* pixels to gray. | [0, 50] |

### C. DCNN Models

In this study, we investigate DCNNs with different architectures and parameters for the best performance on four different dermatoscopic image datasets. As shown in Table II, various DCNNs from the classic VGG series [30] to the latest RegNet series [20] are tested sequentially. The output dimension of the last fully connected layer (FC) of all models is modified to match the number of classes in the corresponding dataset. According to the test results in Table IV, RegNetY-3.2G achieves the best BACC value on ISIC 2018 dataset. Table V indicates that RegNetY-1.6G, RegNetY-8.0G and RegNetY-800M perform best on ISIC 2017，ISIC 2019, and 7-PT Datasets respectively. Since our long-term goal focus on developing single-model-based diagnosis devices, all the subsequent studies in this paper are thereby developed based on the best-performed DCNNs for corresponding datasets.

In order to further prevent overfitting, we redesign a new model, namely RegNetY-##-Drop (## represents unspecified parameters, e.g. "3.3G"), by inserting the DropBlock layer after Stage 3 and Stage 4, and Dropout layer before the last FC layer to the RegNetY model. This design is based on the rationale that DropOut only works in an FC layer and DropBlock works in a convolutional layer [36]. Fig. 3 shows the architecture of RegNetY-##-Drop, where the DropBlock layers share parameter $s$ and all the three regularization modules share parameter $p$. Here parameter $p$ controls how many features to drop and parameter $s$ is the size of the block to be dropped.

### D. Modified RandAugment

Based on RandAugment [39], we propose a modified RandAugment strategy that is more suitable for classification tasks on dermoscopic image datasets. In order to maximize the diversity of samples, this strategy expands the types of transformations used in the search space of RandAugment from 14 to 21, with the intrinsic operations and the corresponding ranges of magnitude listed in Table III. Compared with RandAugment, there are 10 newly added transformations in the search space: Sample_pairing [63], Color_casting [64], Flip, Gauss_noise, Equalize_yuv, Vignetting [64], Scale_diff, Distortion [64], Scale, and Cutout [65]. Among them, the regularization methods such as Sample_pairing, Gauss_noise and Cutout, which have been proven to be effective in preventing overfitting, are deliberately added. Additionally, the transformation of "Identity" that can be achieved by calling a transform with probability set to 0 [39] is not listed in Table III. "TranslateX" and "TranslateY" of RandAugment are not adopted for that their functions are covered by the random cropping performed subsequently.

In addition to the number of transformation $N$ and the augmentation magnitude $M$, the probability of executing each operation also plays an important role in training. Therefore, a probability parameter $P$ is introduced to control whether the selected transformation should be executed or not. In order to avoid the dramatic increase of the search space scale, we share the same $P$ value for all the operations so that each transformation has a probability of $1 - P$ to remain the original image unchanged. In addition, we found that using a random value within the allowable range performs better than using a specific amplitude $M$ value on the HAM dataset, possibly owing to the increased diversity for transformations. Therefore, we finally use only two hyperparameters (i.e., the number of transformation $N$ and the execution probability $P$) to control the data enhancement process in the Modified RandAugment strategy, leaving the amplitude of each transformation randomly selected within the allowable range.

The transformations listed in Table III is classified into 2 categories where color transformations change the color-related properties and shape transformations change the shape-related properties. In RandAugment, each operation is randomly selected from all the transformations without differentiating categories [39]. Therefore, it is likely that all the operations are selected from the same category without touching the other category in the case of $N > 1$. In Modified RandAugment, we hypothesize that equivalent application of operations from both the color and the shape categories to each image will improve the training outcome. Therefore, we divide the transformations in search space into two subsets of {color} and {shape}, comprising 13 and 8 transformations respectively. As $N > 1$, the transformations are randomly selected from {color} and {shape} subsets consequenctly. The working protocol of Modified RandAugment is illustrated below using $N = 2$ as an example:

Step 1. For each image in the training set, randomly select a transformation from {*color*}, and implement the transformation at the preset probability of *P*. If the transformation is to be performed, its amplitude is randomly selected within an allowable range. Otherwise, the original image is remained.

TABLE IV
BACC OF DIFFERENT DCNN MODELS ON THE ISIC 2018 SKIN LESION
CLASSIFICATION CHALLENGE TEST SET.

| model | BACC | model | BACC | model | BACC |
|---|---|---|---|---|---|
| VGG-11 | 0.769 | DenseNet-169 | 0.836 | RegNetX-3.2G | 0.842 |
| VGG-13 | 0.771 | DenseNet-201 | 0.829 | RegNetX-4.0G | 0.834 |
| VGG-16 | 0.745 | DenseNet-161 | 0.837 | RegNetX-8.0G | 0.831 |
| VGG-19 | 0.750 | EfficientNet-b0 | 0.838 | RegNetX-16G | 0.835 |
| ResNet-18 | 0.812 | EfficientNet-b1 | 0.842 | RegNetX-32G | 0.832 |
| ResNet-34 | 0.825 | EfficientNet-b2 | 0.853 | RegNetY-400M | 0.839 |
| ResNet-50 | 0.834 | EfficientNet-b3 | 0.845 | RegNetY-800M | 0.846 |
| ResNet-101 | 0.838 | EfficientNet-b4 | 0.842 | RegNetY-1.6G | 0.850 |
| ResNet-152 | 0.835 | EfficientNet-b5 | 0.843 | RegNetY-3.2G | **0.858** |
| SENet-50 | 0.832 | EfficientNet-b6 | 0.848 | RegNetY-4.0G | 0.848 |
| SENet-101 | 0.845 | EfficientNet-b7 | 0.847 | RegNetY-8.0G | 0.846 |
| SENet-152 | 0.835 | RegNetX-400M | 0.823 | RegNetY-16G | 0.849 |
| SENet-154 | 0.838 | RegNetX-800M | 0.828 | RegNetY-32G | 0.851 |
| DenseNet-121 | 0.832 | RegNetX-1.6G | 0.833 | | |

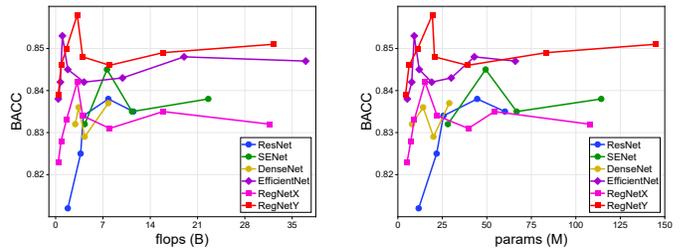

Fig. 4. BACC of different DCNN models on the ISIC 2018 skin lesion classification challenge test set (VGG models are not listed due to their poor performance). Left: Model Compute Capability vs. BACC. Right: Model Size vs. BACC.

Step 2. Randomly select a transformation from {*shape*}, and follow the same procedure in Step 1 to implement the transformation.

Step 3. Randomly crop an image of 224 x 224 from the augmented image obtained from Step 2 and send it to the training network.

### E. Multi-weighted New Loss Method

In the conventional Class-Balanced Loss method [46], the loss weighting factor $w_i$ is calculated by (2) and the weighting strength floats within $[1, 1/N_i)$ regardless of the value of *β*. In our Multi-weighted New Loss method, we modify $w_i$ by adding the hyperparameter *α*, as shown in (7):

$$w_i = (\frac{1}{N_i})^\alpha. \quad (7)$$

Clearly, the scope of $w_i$ is extended beyond $[1, 1/N_i]$ as *α* > 1, which will further enhance the diversity of the weighting strength. In order to reduce the sample imbalance between classes and the classification difficulty imbalance, we propose a Multi-weighted Focal Loss (MWL$_{focal}$) function by combining (7) with the Focal Loss function [45]. Here, a weighting coefficient $C_i$ on the basis is also introduced to strengthen the training effect for specified category, and the MWL$_{focal}$ function is expressed as follow:

$$\text{MWL}_{focal}(z,y) = -C_y(\frac{1}{N_y})^\alpha \sum_{i=1}^{C}(1-p_i^t)^r \log(p_i^t). \quad (8)$$

where $C_y$ is the weighting coefficient for the ground-truth class *y*. For example, we can simply set the category weighting coefficient $C_{MEL}$ a value greater than 1 while keep the other $C_i$ values as 1 to strengthen the training effect for melanoma.

As shown in formula (8), for outliers of $p_i^t \to 0$, the loss→∞, and it would seriously mislead the optimization of network training. We further modify (8) by introducing a correction term to reduce the interference of outliers, and the Multi-weighted New Loss function (MWNL) is updated as:

$$\text{MWNL}(z,y) = -C_y(\frac{1}{N_y})^\alpha \sum_{i=1}^{C} Loss_i. \quad (9)$$

where

$$Loss_i = \begin{cases} (1-p_i^t)^r \log(p_i^t) & p_i^t > T \\ G^* & p_i^t \leq T \end{cases} \quad (10)$$

where *T* is a hyperparameter threshold for limiting the loss of outlier. Experimental results indicate that the value of 0.1 perform best, and *T* is set as 0.1 for all subsequent experiments. $G^* = (1-T)^r \log(T)$ is a constant determined by *T*.

### F. Cumulative Learning Strategy

Inspired by literature [49] and [50], we proposed a novel end-to-end cumulative learning strategy (CLS) and updated (9) as:

$$\text{MWNL}(z,y) = -(C_y^* \frac{1}{N_y})^\beta \sum_{i=1}^{C} Loss_i. \quad (11)$$

where $C_y^{*\alpha} = C_y$, $\beta \in [0, \alpha]$ and changes with the current epoch *E*:

$$\beta = \begin{cases} 0 & E \leq E_1 \\ (\frac{E-E_1}{E_2-E_1})^2 \alpha & E_1 < E < E_2 \\ \alpha & E \geq E_2 \end{cases} \quad (12)$$

where $E_1$ and $E_2$ are hyperparameter thresholds for training epoch, which are set as 20 and 60 respectively in our experiment. With the increase of epoch, *β* changes from 0 to *α*, and the weighted factors for different classes gradually vary from 1 to $C_i(1/N_i)^\alpha$. Wherein, the CLS first trains networks on the originally imbalanced data as usual to make appropriate weight justifications for deep layers' features. Following that the training gradually changes to a re-balancing mode. For more details in the re-balancing training, the learning rate gradually decreases to a small value, coupled with the non-convexity of the loss, only minor changes occurred to the weights of the deep features and the network consequently transfers to the optimization of the upper classifier. Another point to mention here is that the CLS method does not require

TABLE V
DCNNs PERFORM BEST ON 7-PT DATASET, ISIC 2017, AND ISIC 2019.

| dataset | 7-PT | ISIC 2017 | ISIC 2019 |
|---|---|---|---|
| **Best model** | RegNetY-800M | RegNetY-1.6G | RegNetY-8.0G |
| **BACC** | 0.652 | 0.743 | 0.590 |

TABLE VI
BACC OF REGNETY-3.2G WITH ADDING DROPOUT AND DROPBLOCK ON THE ISIC 2018 SKIN LESION CLASSIFICATION CHALLENGE TEST SET.

| $s = 5$ | | | | | |
|---|---|---|---|---|---|
| $p$ | 0.1 | 0.2 | 0.3 | 0.4 | 0.5 |
| BACC | **0.860** | 0.856 | 0.850 | 0.848 | 0.843 |
| $p = 0.1$ | | | | | |
| $s$ | 3 | 4 | 5 | 6 | 7 |
| BACC | 0.855 | 0.857 | **0.860** | 0.859 | 0.852 |

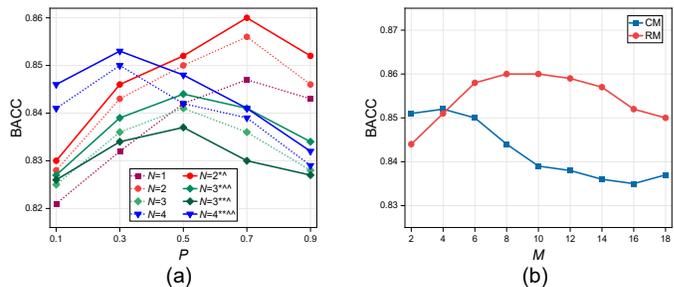

Fig. 5. BACC on the ISIC 2018 test set using Modified RandAugment. (a) BACC with different *N* and *P*. * means randomly select a transformation from the {*color*} subset, ^ means randomly select a transformation from the {*shape*} subset, and transformations are executed following the order of symbols in the training. In addition, transformation is randomly selected from the entire search space for unmarked groups. (b) BACC with the changing of *M* (*N*=2*^, *P*=0.7). CM is "Constant Magnitude", RM is "Random Magnitude with Increasing Upper Bound".

TABLE VII
BACC COMPARISON OF GENERAL AUGMENTATION MATHOD, RANDAUGMENT AND OUR MODIFIED RANDAUGMENT ON THE DIFFERENT TEST SETS.

| Augment method | 7-PT | ISIC 2017 | ISIC 2018 | ISIC 2019 |
|---|---|---|---|---|
| General Method | 0.641 | 0.732 | 0.831 | 0.568 |
| RandAugment | 0.647 | 0.741 | 0.850 | 0.586 |
| **Modified RandAugment** | **0.657** | **0.746** | **0.860** | **0.593** |

additional training time, which is almost half of the BBN method [49].

### G. Training and Evaluation Strategies

The models are initialized with the weights pre-trained on the ImageNet dataset [14], and fine-tuned on the training sets. As shown in Fig. 2, the proposed data enhanced strategy is applied to each image in the training set and the resultant image is randomly cropped in order to match the size of 224 x 224 required by the models.

A multi-crop strategy is adopted for evaluation of the DCNN models. As the indicated by the flowchart in Fig. 2, the areas from the upper left corner to the lower right corner of the test image are cropped with equal space, and the average of their predicted values is used as the final prediction value. The number of crops is defined as 16 in our study since we notice little improvement of the model performance beyond this number.

All the training and the testing tasks are performed on 2 NVIDIA GeForce GTX 2080Ti graphics cards using PyTorch. Adam [66] is used as an optimizer and "MultiStepLR" is used for learning rate schedule for all the models. A starting learning rate of 0.001 is chosen and is gradually reduced at a factor of $\lambda = 0.1$ per 10 epochs from the 30th epochs. In the case that the training loss cannot be reduced to a smaller value during training, a starting learning rate of 0.0004, 0.0002 or 0.0001 is chosen. In order to prevent overfitting, we adopt an early stopping strategy to stop optimization after 70 epochs. We also tried other regularization methods combating overfitting, such as label smoothing [67], parameter norm penalties [68] and the optimizers of SGD [68], RMSProp [69], and Nadam [70], but they were not adopted due to poor performance. Most of the models in this study use a batch size of 128. However, the batch size is reduced for some of the large models in order to match the memory capacity of the graphic cards since these models require large memory due to their feature map sizes.

## IV. EXPERIMENTS AND RESULTS

### A. Performance of different DCNN Models

First, different DCNNs are implemented for training and testing tasks on the ISIC 2018 skin lesion classification challenge dataset. The proposed Modified RandAugment strategy and Cross Entropy Loss re-weighted by (1) are adopted in the training. Table IV and Fig. 4 show the performances of DCNNs in different architectures and capacities. VGG models are not listed in Fig. 4 due to their poor performance. For the models of similar architectures, such as those from Resnets to Regnets, the BACC values typically increases to a maximum value and then decreases as the capacity increases. This observation implies that, for a dataset with insufficient samples such as the HAM dataset used in this study, only the model with moderate complexity yields the best performance.

According to Table IV and Fig. 4, the most recently proposed models (e.g. EfficientNet and RegNetY) that have achieved excellent performance in general classification tasks (e.g. ImageNet) also show better performance on the ISIC2018 challenge test set. This observation implies that the DCNNs with more advanced architectures can achieve better performance than traditional ones in medical imaging tasks such as skin lesion classification. We also notice that RegNetYs generally perform better than EfficientNets. RegNetY-3.2G achieves the best 0.858 BACC, which is 0.005 higher than the best result of EfficientNet-b2. Meanwhile, RegNetYs with additional SE modules has better result than RegNetXs. The improvement in BACC validates that the addition of SE structure in DCNNs is effective for not only natural image but also dermoscopic classification tasks.

Similarly, we execute the test on the other three datasets, and their corresponding best performance are shown in Table V. For the ISIC 2019 dataset of 25331 training samples, the best network is RegNetY-8.0G. For the smaller ISIC 2017 dataset, the best network is RegNetY-1.6G, and the best network for the smallest 7-PT Dataset is RegNetY-800M. Obviously, the capacities of best-performed DCNNs for various datasets are different and positively correlated with their data size, which further prove our above conclusions. Finally, the four DCNNs have the best performance on the corresponding dataset are selected as baselines for all subsequent experiments.

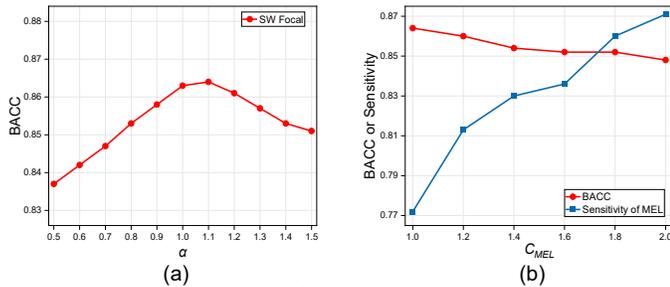

Fig. 6. Performance on the ISIC 2018 test set using our proposed $MWL_{focal}$ method. (a) BACC of $MWL_{focal}(C_{MEL}=1.0)$ with different $\alpha$. (b) BACC and melanoma sensitivity of $MWL_{focal}(\alpha=1.1)$ with different $C_{MEL}$.

TABLE VIII
BACC COMPARISON OF DIFFERENT LOSS FUNCTIONS OR DIFFERENT TRAINING STRATEGIES ON THE DIFFERENT TEST SETS.

| Approach | 7-PT | ISIC 2017 | ISIC 2018 | ISIC 2019 |
|---|---|---|---|---|
| CE | 0.598 | 0.729 | 0.831 | 0.556 |
| CE-RS | 0.657 | 0.739 | 0.849 | 0.570 |
| CE-RW | 0.657 | 0.746 | 0.860 | 0.593 |
| $CB_{focal}$ [46] | 0.674 | 0.764 | 0.862 | 0.599 |
| LDAM [50] | 0.677 | 0.763 | 0.865 | 0.611 |
| LOW [71] | 0.672 | 0.757 | 0.853 | 0.601 |
| CCE [72] | 0.667 | 0.748 | 0.848 | 0.564 |
| $MWL_{focal}$ | 0.678 | 0.766 | 0.864 | 0.606 |
| MWNL | **0.688** | **0.775** | 0.867 | 0.614 |
| BBN [49] | 0.630 | 0.771 | 0.861 | 0.622 |
| LDAM-DRW [50] | 0.603 | 0.765 | 0.863 | 0.626 |
| MWNL-DRW | 0.610 | 0.769 | 0.862 | 0.629 |
| LDAM-CLS | 0.608 | 0.762 | 0.873 | 0.639 |
| MWNL-CLS | 0.613 | 0.763 | **0.875** | **0.644** |

As shown in Fig. 3, the DropOut and DropBlock layers are added to the RegNetY network as the regularization modules. Table VI shows the results on the ISIC 2018 test set. According to the table, BACC achieves the best value of 0.860 at $p=0.1$ and $s=5$, which is 0.002 higher than that of original model. In contrast, the BACC values drop below that of original RegNetY-3.2G with the further increase of $p$. The above operations obtained similar experimental performance on the other three datasets.

### B. Results on Modified RandAugment

Our proposed augmentation method is mainly validated using RegNetY-3.2G-Drop as baseline and Cross Entropy Loss re-weighted by (1) as the loss function on the ISIC 2018 test set. Fig. 5 (a) shows the resultant BACCs after adopting the Modified RandAugment strategy with different $N$ and $P$ values. Considering the limitations in computing resources and the previous experience in RandAugment where the best $N$ value is usually less than 4 [39], we perform the training and the testing tasks at $N$ values ranging from 1 to 4. According Fig. 5 (a), dividing the transformations into {*color*} and {*shape*} subsets and randomly selecting transformations from the alternate subsets result in BACC values greater than randomly selecting and orderly executing transformation from the entire search space. One exception is observed for $N=3**^\wedge$, where the BACC value is lower than that at $N=3$.

Our experimental results show that the change of execution probability $P$ has a significant impact on BACC. Taking $N=2*^\wedge$ as an example, BACC is only 0.832 at $P=0.1$ but reaches the best value of 0.860 at $P=0.7$. Our experimental results also indicate that keeping the ratio of transformed samples in a proper range during training helps to improve performance. For example, the best BACC occurs at $P=0.7$ when $N=1$ or 2, while this optimal value is achieved at $P=0.3$ when $N=4$. The reason may be that a low ratio is not conducive to increase the data diversity because more samples remain unchanged, while a high ratio meaning more transformed samples may excessively change the inherent characteristics distribution of the original samples.

In comparison with Modified RandAugment, RandAugment uses the parameter $M$ (a single global distortion magnitude for all transformations) instead of $P$ to optimize BACC [39]. In order to evaluate the effects of the $M$ value on the BACC performance, we split the deformation ranges of the transformations into 10 equally spaced levels except those uninvolved with magnitude. When $M=10$, the transformation magnitude reaches the corresponding maximum value in Table III, and it will exceed their maximum value at $M>10$. The effect of $M$ is verified following two different schemes of "Constant Magnitude (CM)" and "Random Magnitude with Increasing Upper Bound (RM)". CM refers to a constant value for each transformation magnitude; while RM refers to a randomly selected value between 0 and $M$ for each transformation magnitude. Fig. 5 (b) shows the BACC performance of RegNetY-3.2G-Drop after adopting the Modified RandAugment strategy with different $M$ values at $N=2*^\wedge$ and $P=0.7$. According to the figure, the BACC performance of RM is superior to that of CM except at $M=2$ or 4, indicating that a randomly distorted transformation magnitude expands the diversity of sample transformations and improves the BACC performance. As for RM, the BACC value fluctuates insignificantly when $M$ changes within the interval of [6,14], and reaches the maximum at $M=10$, indicating that the augment effect is insensitive to the transformation magnitude within a certain range. Therefore, $M$ is set as 10 to reduce the search space for all experiments involved in the Modified RandAugment strategy.

Table VII shows the best BACC performance obtained by Modified RandAugment, RandAugment and general augmentation method on 4 different test sets. Here, the general augmentation method consists of transformations like random flipping of images, random change of brightness, contrast, saturation and hue, and random cropping. Clearly, both RandAugment and Modified RandAugment greatly improve the performance, superior to the general augmentation method. In comparison with RandAugment, Modified RandAugment further improves the value of BACC, verifying its effectiveness for the dermatoscopic image datasets.

### C. Results on Multi-weighted New Loss

First, the effectiveness of our proposed $MWL_{focal}$ function is tested using RegNetY-3.2G-Drop as the baseline and Modified RandAugment as the augmentation method on the ISIC 2018 test set. Fig. 6 (a) shows the BACC values after adopting MWL$_{focal}$ functions at different $\alpha$ levels (here all of $C_i$ are set to 1.0 and $r$ is fixed as 2.0). As $\alpha$ increases, the value of BACC also increases until it reaches the maximum of 0.864 at $\alpha$=1.1. After that, further increase of $\alpha$ gradually reduces the BACC level. Noticeably, as an extension of the CB$_{focal}$ [46] in weighting strength, MWL$_{focal}$ ($C_{MEL}$=1.0) at $\alpha$=1.0 yields a BACC value of 0.862, corresponding to the best performance achievable by CB$_{focal}$. This result verifies that MWL$_{focal}$ is better than CB$_{focal}$ in performance improvement.

TABLE IX
THE PERFORMANCE OF ISIC 2018 CHALLENGE WINNERS FROM THE LEGACY LEADERBOARD (ROWS 1–3), CURRENT ISIC 2018 CHALLENGE LIVE LEADERBOARD (ROWS 4–8), AND OUR PROPOSED APPROACH.

| Team / authors | Use external data | Use ensemble models | Sen. MEL | Avg. Spec | Avg. AUC | BACC |
|---|---|---|---|---|---|---|
| # 1 | Yes | Yes | 0.760 | 0.833 | 0.983 | 0.885 |
| # 2 [44] | Yes | Yes | 0.801 | 0.984 | 0.987 | 0.856 |
| # 3 | No | Yes | 0.702 | 0.980 | 0.978 | 0.845 |
| ## 1 | Yes | Yes | 0.778 | 0.981 | 0.982 | **0.895** |
| ## 2 | No | - | 0.813 | 0.975 | 0.983 | 0.886 |
| ## 3 [55] | Yes | Yes | 0.585 | 0.992 | 0.979 | 0.874 |
| ## 4 | No | - | 0.860 | 0.980 | 0.953 | 0.873 |
| ## 5 | Yes | Yes | 0.830 | 0.976 | 0.976 | 0.866 |
| Our MWNL | No | No | 0.784 | 0.975 | 0.979 | 0.867 |
| Our MWNL-CLS | No | No | 0.819 | 0.980 | 0.985 | 0.875 |

TABLE X
THE PERFORMANCE OF ISIC 2019 CHALLENGE WINNERS FROM THE LEGACY LEADERBOARD (ROWS 1–3), CURRENT ISIC 2019 CHALLENGE LIVE LEADERBOARD (ROWS 4–8), AND OUR PROPOSED APPROACH.

| Team / authors | Use external data | Use ensemble models | Sen. MEL | Avg. Spec | Avg. AUC | BACC |
|---|---|---|---|---|---|---|
| # 1 [62] | Yes | Yes | 0.594 | 0.977 | 0.923 | 0.636 |
| # 2 | No | Yes | 0.675 | 0.952 | 0.780 | 0.607 |
| # 3 | No | Yes | 0.684 | 0.963 | 0.886 | 0.593 |
| ## 1 | Yes | Yes | 0.778 | 0.981 | 0.982 | **0.662** |
| ## 2 | Yes | Yes | 0.813 | 0.975 | 0.983 | 0.648 |
| ## 3 | Yes | Yes | 0.585 | 0.992 | 0.979 | 0.641 |
| ## 4 [73] | Yes | Yes | 0.860 | 0.980 | 0.953 | 0.640 |
| ## 5 | No | - | 0.830 | 0.976 | 0.976 | 0.640 |
| Our MWNL | No | No | 0.684 | 0.948 | 0.872 | 0.614 |
| Our MWNL-CLS | No | No | 0.712 | 0.951 | 0.868 | 0.645 |

TABLE XI
THE PERFORMANCE OF THREE TOP-RANKING CHALLENGE SOLUTIONS (ROWS 1–3), FOUR RECENT METHODS (ROWS 4–7), AND OUR PROPOSED APPROACH ON THE ISIC 2017 TEST SET.

| Team / authors | Use external data | Use ensemble models | Sen. MEL | Avg. Spec [a] | BACC | Avg. AUC [b] |
|---|---|---|---|---|---|---|
| #1 [74] | Yes | Yes | 0.735 | 0.812 | 0.831 | 0.911 |
| #2 [75] | Yes | No | 0.103 | 0.998 | 0.883 | 0.910 |
| #3 [51] | Yes | Yes | 0.547 | 0.970 | 0.844 | 0.908 |
| Xie et al. [10] | Yes | Yes | 0.727 | 0.930 | - | **0.938** |
| Zhang et al. [76] | Yes | No | 0.658 | 0.882 | - | 0.917 |
| Xie et al. [77] | Yes | No | 0.556 | 0.910 | - | 0.916 |
| Zhang et al. [78] | Yes | No | - | - | - | 0.913 |
| Our MWNL | No | No | 0.607 | 0.680 | 0.775 | 0.923 |
| Our MWNL-CLS | No | No | 0.564 | 0.760 | 0.763 | 0.917 |

[a] Avg. Spec: the average specificity of MEL and SK.
[b] Avg. AUC: the average AUC of MEL and SK. Here, Avg. AUC is the official key metric for ISIC 2017 classification challenge.

TABLE XII
THE PERFORMANCE OF THREE RECENT METHODS (ROWS 1–3), AND OUR PROPOSED APPROACH ON THE 7-PT DATASET.

| Team / authors | Use external data | Use ensemble models | Sen. MEL | Avg. specificity | Avg. AUC | BACC |
|---|---|---|---|---|---|---|
| Nedelcu et al. [79] | Yes | Yes | 0.673 | 0.926 | - | 0.638 |
| Kawahara et al. [57] | Yes | Yes | 0.614 | 0.910 | 0.896 | 0.604 |
| Rodrigues et al. [80] | Yes | Yes | - | 0.710 | 0.620 | 0.408 |
| **Our MWNL** | No | No | 0.624 | 0.630 | 0.911 | **0.688** |
| Our MWNL-CLS | No | No | 0.607 | 0.647 | 0.912 | 0.613 |

Fig. 6 (b) shows the BACC and melanoma detection sensitivity of models adopting MWL$_{focal}$ of different $C_{MEL}$. At $C_{MEL}$=1.0, the maximal BACC reaches 0.864, while the melanoma detection sensitivity is only 0.772. As $C_{MEL}$ increases, the BACC decreases slightly, while the melanoma detection sensitivity increases significantly. For example, as $C_{MEL}$=2.0, BACC drops to 0.848, and melanoma sensitivity rises up to 0.871. This proves that training a DCNN with both high BACC and high sensitivity of specified class is possible by adjusting $C_i$.

The effectiveness of the proposed MWNL method is further verified on different datasets using RegNetY-##-Drop as the baseline and Modified RandAugment as the augmentation method. Table VIII shows the result of our MWNL with MWL$_{focal}$, the standard training, and several state-of-the-art techniques widely adopted in mitigating data imbalance. Here these methods include: standard cross-entropy loss (CE), CE with over-sample method (CE-RS), CE re-weighted by (1) (CE-RW), CB$_{focal}$ [46], label-distribution-aware margin loss (LDAM) [50], learning optimal samples weights method (LOW) [71], and complement cross entropy (CCE) [72]. It is clear that our MWL$_{focal}$ helps to achieve better performance than those methods except for LDAM, and our MWNL further improved the performance beyond all of them. In view of this, the addition of correction items to suppress the adverse effects of outliers can really improve the performance of DCNNs on small and imbalanced dermoscopic image datasets.

### D. Results on Our Cumulative Learning Strategy

Table VIII also shows the performance of DCNNs trained following our proposed end-to-end Cumulative Learning Strategy (CLS), BBN method [49], and two-stage DRW method [50]. On the ISIC 2018 and ISIC 2019 datasets, the CLS method performs better than other methods, and the MWNL-CLS method achieves the best performance, which proves the effectiveness of the proposed training strategy. However, on the 7-PT and ISIC 2017 datasets, the CLS methods doesn't work well. Especially, BACC dropped significantly on the 7-PT dataset with fewest samples. It is likely that too few samples unsuccessfully enable the effective learning of deep features in the training strategies like DRW and CLS, and thereby fail the proper weights initialization for model in the initial training phase, and finally hinder subsequent optimization training.

It is also noticed that BBN method that works on general imbalanced datasets is not efficient on the four dermatoscopic image datasets. Although the BBN method performs slightly better than the DRW method and our CLS method on 7-PT and ISIC 2017 datasets both of smaller sample sizes, its training time is almost twice that of the other two methods.

### E. Comparison with Other Methods in the Challenge

Table IX - XII show the performance of our method and other state-of-the-art methods on 4 different skin lesion classification tasks at the time of submitting the manuscript.

As shown in Table IX, on the ISIC 2018 classification challenge, our proposed MWNL-CLS approach ranks third overall among the top 200 submissions on the live leaderboard at the time this manuscript is submitted (leaderboard data as of 10/21/2021), and ranks the first among those using a single model without additional data. On the ISIC 2019 classification challenge (Table X), our proposed MWNL-CLS approach also ranks third on the live leaderboard at the time this manuscript is submitted, and surpasses the first-ranked team on the legacy leaderboard, also ranks the first among those using a single model without additional data. As the Table IX and Table X indicate that, the MWNL-CLS method significantly improves all the BACC, Avg. Spec and Sen. MEL than the MWNL method, which further proves the effectiveness of our CLS training strategy on imbalanced datasets that are not very small.

As shown in Table XI, on the ISIC 2017 classification challenge, our proposed MWNL surpasses all the methods on the legacy leaderboard. Compared with other state-of-the-art methods, the MWNL only lower than a method using both additional data and ensemble models [10]. Data in Table XII indicates that our MWNL method is better than all other published ones on the 7-PT dataset. It is worth noting that in our approach, only dermoscopic image data from the 7-PT dataset is used. While in the comparison methods, clinical image data, patient meta data or external data from other datasets are additionally used.

In general, DCNNs with ensemble models perform better than those with a single model [10, 44, 55, 74, 79] , and similar phenomenon can also be observed from Table IX - XII. However, implementing ensemble models is less practical due to the limitation in computing resources and computing time. For example, one of the best methods on the ISIC 2018 challenge live leaderboard integrates 90 DCNNs, which takes 13.9 seconds in a single inference [55]. In comparison, our method takes only 0.05 seconds in the similar computing environment, and achieves better performance. In addition, although our method uses multi-crop strategy in the inference process, the single-model method in asynchronous pipeline mode results in the overall inference time just slightly longer than that adopt single crop method, because these different crops can be calculated in different layers of the model at the same time. Therefore, our method has great potential for deployment on portable diagnostic systems.

## V. CONCLUSION

This paper proposes a novel skin lesion classification method consisting of modified DCNNs integrating regularization DropOut and DropBlock, Modified RandAugment, Multi-Weighted New Loss and an end-to-end Cumulative Learning Strategy. DCNNs of different structures and capacities trained on different dermoscopic image datasets show that the models with moderate complexity outperform the larger ones. The Modified RandAugment helps to achieve significant performance with less computing resources and shorter time. The Multi-weighted New Loss can not only deal with the class imbalance issue, improve the accuracy of key classes, but also reduce the interference of outliers in the network training. The end-to-end cumulative learning strategy can more effectively balance representation learning and classifier learning without additional computational cost. By combining Modified RandAugment and Multi-weighted New Loss, we train single-model DCNNs on ISIC 2018, ISIC 2017, ISIC 2019 and 7-PT Dataset following the end-to-end cumulative learning strategy. They all achieve outstanding classification accuracy on these datasets, matching or even surpassing those ensembling methods. Our study shows that this method is able to achieve a high classification performance at a low cost of computational resources on a small and imbalanced dataset. It is of great potential to explore mobile devices for automated screening of skin lesions and can be also implemented in developing automatic diagnosis tools in other clinical disciplines.